\documentclass[sn-basic]{sn-jnl}

\usepackage{graphicx}%
\usepackage{multirow}%
\usepackage{amsmath,amssymb,amsfonts}%
\usepackage{amsthm}%
\usepackage{mathrsfs}%
\usepackage[title]{appendix}%
\usepackage{xcolor}%
\usepackage{textcomp}%
\usepackage{manyfoot}%
\usepackage{booktabs}%
\usepackage{algorithm}%
\usepackage{algorithmicx}%
\usepackage{algpseudocode}%
\usepackage{listings}%
\usepackage{caption}%
\usepackage{array}%

\theoremstyle{thmstyleone}%
\theoremstyle{thmstyletwo}%

\theoremstyle{thmstylethree}%

\begin{document}

\title{\textbf{Digital Epidemiology: Leveraging Social Media for Insight into Epilepsy and Mental Health}}


\author*[1]{\fnm{Liza} \sur{Dahiya}}\email{lizadahiya@cse.iitb.ac.in}

\author[1]{\fnm{Rachit} \sur{Bagga}}\email{rachitbagga@cse.iitb.ac.in}
\equalcont{These authors contributed equally to this work.}

\affil*[1]{\orgdiv{Computer Science and Engineering}, \orgname{Indian Institute of Technology, Bombay}, \orgaddress{\street{Powai}, \city{Mumbai}, \postcode{400076}, \state{Maharashtra}, \country{India}}}



\abstract{Social media platforms, particularly Reddit's \texttt{r/Epilepsy} community, offer a unique perspective into the experiences of individuals with epilepsy (PWE) and their caregivers. This study analyzes 57k posts and 533k comments to explore key themes across demographics such as age, gender, and relationships. Our findings highlight significant discussions on epilepsy-related challenges, including depression (with 39.75\% of posts indicating severe symptoms), driving restrictions, workplace concerns, and pregnancy-related issues in women with epilepsy. We introduce a novel engagement metric, F(P), which incorporates post length, sentiment scores, and readability to quantify community interaction. This analysis underscores the importance of integrated care addressing both neurological and mental health challenges faced by PWE. The insights from this study inform strategies for targeted support and awareness interventions. The dataset and code are available at \url{https://shorturl.at/neGz7}\footnote{will be replaced by GitHub link upon acceptance.}.}




\maketitle

\section{Introduction}\label{sec1}
Epilepsy is a chronic neurological disorder characterized by recurrent, unprovoked seizures resulting from abnormal electrical activity in the brain. Affecting over 50 million people worldwide, epilepsy presents a significant public health challenge due to its impact on individuals' quality of life \cite{Keezer2016}, social stigmatization, and the need for ongoing medical management.  The causes of epilepsy are diverse, including genetics, brain injuries, infections, and developmental disorders.

According to the World Health Organization \footnote{\url{https://www.who.int/news-room/fact-sheets/detail/epilepsy}}, around 80\% of epilepsy cases occur in low to middle-income countries, where up to 70\% could achieve freedom from seizures with adequate care. Epilepsy triples the risk of early death. In regions with low income, a staggering 75\% number of people do not have access to the necessary treatment, while stigma and discrimination continue to affect numerous people with epilepsy on a global scale.


In recent years, the growth of social media platforms has provided a unique opportunity to observe and analyze public discourse around health issues, including epilepsy. Social media posts offer real-time insights into the experiences, challenges, and support systems of individuals with epilepsy and their caregivers \cite{conway2016social}. The prevalence of social media usage among diverse populations makes it a valuable tool for epidemiological studies, enabling one to gather large datasets that reflect the nuanced experiences of living with epilepsy \cite{guntuku2017detecting}.

The importance of studying social media posts in the context of epilepsy cannot be overstated. These platforms serve as virtual communities where individuals share personal stories, seek advice, and offer support \cite{reavley2014use}. By analyzing the content and patterns of social media posts, researchers can identify common concerns, misconceptions, and general sentiment surrounding epilepsy \cite{joseph2015schizophrenia}. This data can inform public health initiatives, enhance patient education, and improve support networks. Furthermore, analysis of social media posts can help detect emerging trends and topics of interest, providing a dynamic and up-to-date understanding of the impact of epilepsy on individuals' lives \cite{cavazos-rehg2014characterizing}.

\begin{table}[!htp]
    \centering
    \small
    \begin{tabular}{|c|p{6cm}|} 
        \hline
        \textbf{Category} & \textbf{Post Content} \\ \hline \hline
         Question & Does anyone else notice more cherry angiomas or an increase in size after tonic clonic seizures? \\ \hline
         Rant & Shingles vaccine is kicking my ass today! I'm laying in bed watching Charmed. \\ \hline
         Support & Is it because of the disorder or the meds, or am i just balding at 19 for no reason \\ \hline
    \end{tabular}
    \vspace{2pt}
    \captionsetup{font=footnotesize}
    \caption{Examples of post content for different categories. The complete dataset contains more columns like time, author, comments, etc.}
    \label{tab:sample_dataset}
\end{table}

Our study aims to evaluate the content and volume of epilepsy-related social media posts. employing advanced techniques to detect depression levels, analyze the co-occurrence of symptoms, and examine the correlation between specific symptoms and depression severity. By analyzing a large dataset of 57k posts and 553k comments, we strengthen the reliability of our findings, offering a deeper understanding of the concerns and experiences of the epilepsy community. Additionally, our detailed visualizations reveal trends and relationships, shedding light on the themes across different post types, which enhances the scope and depth of our analysis.Through our analyses, we aim to address the following research questions, exploring the demographic, behavioral, and thematic dimensions of experiences shared in the Reddit r/Epilepsy community:

\begin{itemize} 
\item \textbf{RQ1:} Which demographic groups (e.g., age, gender, relationship to people with epilepsy) are most frequently represented in epilepsy-related social media posts, and how do discussion topics vary across these groups? 
\item \textbf{RQ2:} What temporal trends emerge in the dataset, and what insights do they reveal about the evolving experiences and concerns of people with epilepsy over time? 
\item \textbf{RQ3:} How can depression signals be effectively detected in epilepsy-related posts, and how do these signals differ across demographic groups? Furthermore, how does user engagement vary between posts that show signs of depression and those that do not? 
\end{itemize}

Our study makes several key contributions. First, our demographic analysis revealed notable differences in discussion topics across age, gender, and relationship status, emphasizing the necessity for personalized support strategies. Second, we analyzed the co-occurrence of symptoms and explored the discussions around medications, offering valuable insights into the shared challenges and treatment approaches faced by the epilepsy community. Third, we conducted a temporal analysis of the r/Epilepsy subreddit, uncovering significant growth in user engagement over time, particularly around epilepsy awareness events and key dates, which underscores the platform’s role as a hub for outreach and support. Fourth, we introduced a novel engagement metric, F(P), which integrates post length, sentiment analysis, and readability scores, providing a more nuanced understanding of community interaction and post engagement. Lastly, leveraging the DepRoBERTa-large-depression model, we detected a high prevalence of depression among users, shedding light on the emotional burden of epilepsy and the critical need for targeted mental health interventions. 
\section{Related Work}\label{sec2}

\begin{figure*}[!htp]
    \centering
    \includegraphics[width=0.9\linewidth]{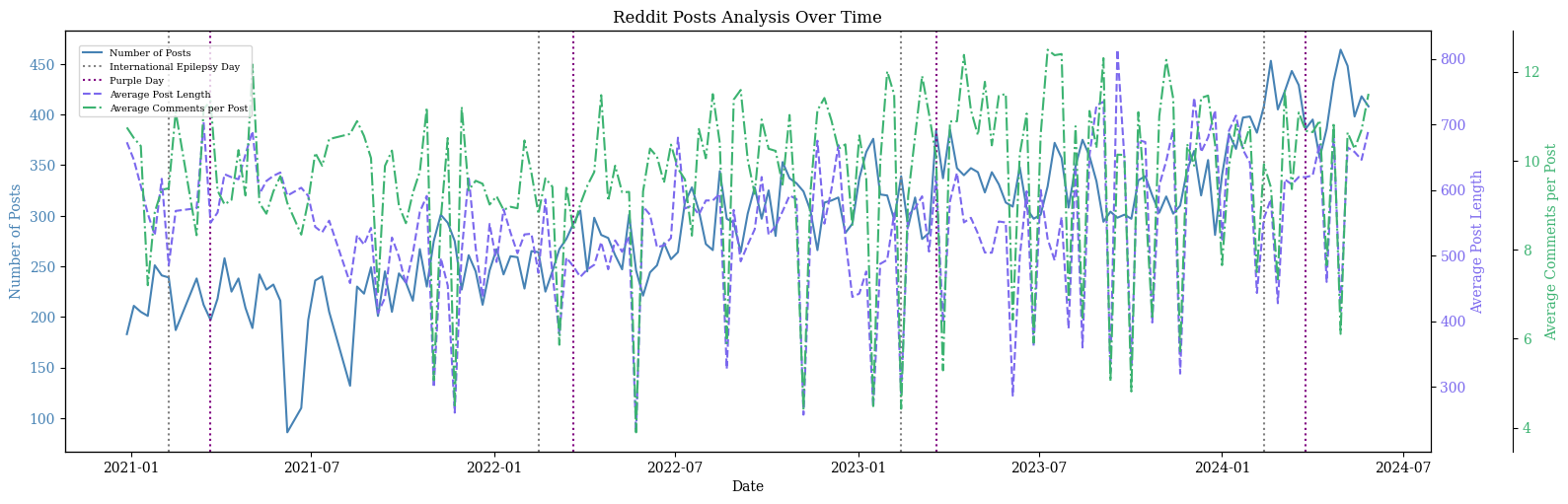}
    \caption{Temporal Distribution of r/Epilepsy posts.}
    \label{fig:temporalGraph}
\end{figure*}
Social media platforms have become integral for studying public discourse and community dynamics related to mental health and neurological disorders. Various studies \cite{Rivas2020, To2023, DomingoEspiñeira2024} have utilized platforms like Twitter, Facebook to analyze health-related conversations, offering insights into how different demographic groups engage with topics surrounding mental health and neurological conditions.

For Epilepsy, online support groups, such as those hosted by the Epilepsy Foundation\footnote{\url{https://www.epilepsy.com/}} and Mayo Clinic\footnote{\url{https://www.mayoclinic.org/}}, play a pivotal role in facilitating discussions on complex issues. Research underscores that these forums enable profound conversations and interventions beyond conventional medical settings \cite{HE201965, Dueweke2019}. Studies highlight the significant stigma associated with epilepsy, emphasizing the necessity for cultural sensitivity in discussions \cite{MAYOR2022142}. Falcone et. all~\cite{Falcone2020} conducted a seminal study examining suicide-related conversations among epilepsy patients, specifically studying difference in conversations styles and themes among teens vs. adults. However, the study was limited to data collected from US IP addresses in a single year.

Social media platforms like Twitter and Facebook also provide valuable data for understanding epilepsy-related discussions. A study by Meng~\cite{Meng2017} revealed a 100\% increase in epilepsy-related posts from 2012 to 2016, indicating heightened engagement levels and awareness. Thematic analyses~\cite{barbara2021} of these conversations have highlighted key issues faced by people with epilepsy (PWE), such as disease awareness, psychological and physical impacts of seizures, the importance of sleep, treatment efficacy, and mental health challenges. Another study~\cite{Revson2016} explored the multilingual perception of epilepsy on Twitter, identifying significant stigma associated with terms like ``seizure" compared to ``epilepsy," revealing variations in public perception across languages.

While these studies underscore the complexity and depth of social media as a field for understanding epilepsy and its social ramifications, several limitations exist. Notably, many studies are not generalizable due to small dataset sizes and lack of demographic analysis \cite{Meng2017, Revson2016,popoola2023instagram}. Results could differ significantly across demographics, but most research does not account for this variability. Additionally, some studies analyze restrictive datasets belonging to one demographic group \cite{alsalem2021epilepsy, panagariya2019prevalence}, further limiting their applicability.

Another critical gap in the existing literature is the intersection of epilepsy and depression. Depression is very common among epilepsy patients \cite{jackson2005depression, kanner2002depression}, and some studies \cite{gilliam2006rapid, hansen2015combined, friedman2009identifying} have tried to detect it clinically. However, the use of social media content for detecting depression in epilepsy patients has not been explored. While social media has been utilized for depression detection generally \cite{lin2020sensemood, william2021text, zeberga2022retracted} and for some other metal health diseases such as stroke patients \cite{schubert1992detection, sagen2010early} and patients suffering from chronic physical diseases \cite{goldberg2010detection}, it remains underexplored for epilepsy patients. This gap is significant given the potential for social media to provide real-time insights into the mental health status of individuals.

Our study addresses these gaps by: 1) Analyzing over 57,681 posts and 533,994 comments from the \texttt{r/Epilepsy} subreddit, 2) Leveraging a larger and more diverse dataset compared to platforms like Twitter \cite{Meng2017, Revson2016}, where posts are length-restricted. Reddit also offers a unique perspective with its structure encouraging detailed posts and in-depth, judgment-free discussions, 3) Providing a comprehensive examination of epilepsy-related discussions and performing demographic analysis, reflecting how results can change across different groups, and 4) Utilizing social media content to detect depression signals and examining how the community reacts to these posts.
    

\section{Dataset}\label{sec3}

\subsection{Data Preparation}\label{subsec2}
Data for this study was collected from the social media platform Reddit \footnote{\url{https://www.reddit.com/}}, known for its diverse user-generated content and discussion forums. Specifically, we collected data from the subreddit \texttt{r/Epilepsy}, which has 45K members and receives more than 40 daily posts.
PullPush.io \footnote{\url{https://pullpush.io/}} is an open-source service for the indexing and retrieval of content that Reddit users have submitted to Reddit. We used this API to fetch about 57,681 posts along with all their comments (a total of 533,994 comments) from the subreddit \texttt{r/Epilepsy}. This dataset spans over 3 years, from May 2021 to May 2024.

\subsection{Preliminary Data Statistics}\label{subsec3}
To gain a deeper understanding of community engagement and content themes, we analyzed several key metrics. These metrics, presented in Table~\ref{tab:statistics}, include average length of post content, average length of post title, average number of comments received on posts, average posts per author, average number of daily posts, most common post type, and most common relationship mentioned. This analysis highlights the interactive nature of the forum, indicating a high level of community support and discussion.
\begin{table}[h!]
    \small
    \centering
    \renewcommand{\arraystretch}{1} 
    \begin{tabular}{|l|r|}
        \hline
        \textbf{Metric} & \textbf{Statistic} \\ \hline \hline
        Avg. length of post content & 528.89 \\ \hline
        Avg. length of post title & 39.91 \\ \hline
        Avg. number of comments & 9.257 \\ \hline
        Avg. length of comments & 286.753 \\ \hline
        Avg. posts per author & 3.169 \\ \hline
        Avg. daily posts & 45.88 \\ \hline
        Most common post type & Question \\ \hline
        Most common relationship (apart from self) & Children \\ \hline
    \end{tabular}
    \captionsetup{font=small}
    \caption{Statistical Analysis of Epilepsy Subreddit Posts}
    \label{tab:statistics}
\end{table}

\section{Methods}\label{sec4}
\subsection{RQ1: Demographic Analysis}\label{subsec4}
\begin{figure}[!htp]
    \centering
    \includegraphics[width=0.8\linewidth]{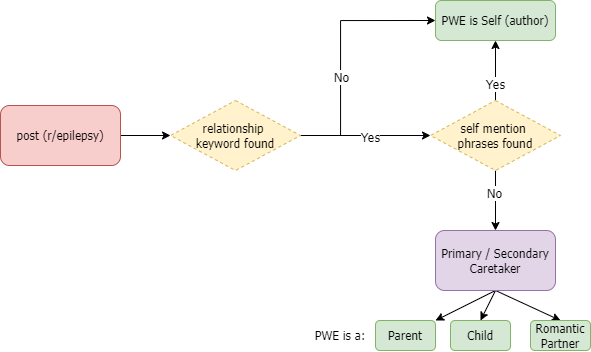}
    \caption{Relationship (to PWE) extraction pipeline.}
    \label{fig:relationship_extraction}
\end{figure}

\paragraph{Who is the PWE?} 
The subreddit \texttt{r/Epilepsy} features posts written either by individuals with epilepsy (PWE) or by their primary and secondary caregivers. These posts serve various purposes, including asking questions (46.9\% of total posts), sharing personal stories (support posts make up 9\% of total posts), or expressing frustration (rants account for 10\% of total posts). To determine whether the post author is a PWE or their primary/secondary caregiver, we developed a pipeline to classify the text into `self' (PWE) and caregiver categories, such as `romantic' (e.g., spouse or partner), `parents,' and `children.' This pipeline is illustrated in Figure~\ref{fig:relationship_extraction}.

The relationship classification relies on specific keywords and phrases. For instance, posts containing terms like ``boyfriend", ``girlfriend", ``husband", and ``wife" are classified as `romantic', while terms such as ``mom", ``dad", or ``guardian" fall under `parents'. Similarly, posts mentioning ``son", ``daughter", or ``toddler" are categorized under 'children.' Additionally, PWEs sometimes refer to relationships with their parents or spouses to thank them for their support or express concerns about disappointing them. In such cases, further phrases like ``I have been diagnosed with epilepsy," ``my seizures," or ``I have epilepsy" are used to classify the post as `self’. To prevent misclassification, a diverse set of similar expressions is incorporated. The methodology includes a comprehensive set of patterns to ensure accurate post categorization and reduce the risk of errors. To assess the statistical significance of the relationship-based groups, we conducted Chi-Square tests and pairwise t-tests, utilizing the top $K=50$ TF-IDF features for each of the four relationship categories.


To further explore the specific concerns and topics discussed within these relationship categories — how they differ in terms of what individuals talk about depending on their relationship to the PWE — we employed Topic Modeling using Latent Dirichlet Allocation (LDA) \cite{blei2003latent}. After preprocessing the text by removing stopwords (including common epilepsy-related terms such as ``seizure", ``keppra"), and lemmatizing, we tokenized the data and applied LDA with a total of $K=3$ topics for each relationship category. To identify both salient and relevant terms, we experimented with different relevance settings for $\lambda$ \cite{sievert2014ldavis}. This allowed us to capture the most important and distinctive terms in each topic, revealing patterns in how PWE and their caregivers discuss their experiences across various relational contexts.

\paragraph{Average age of PWE}
Epilepsy affects individuals of all ages, from children to adults and the elderly. The relationship between epilepsy and age has been extensively studied over the years~\cite{hauser1992seizure, beghi2018aging, tang2015prevalence}. In this study, we aim to identify which age groups are mentioned in social media posts, analyze the specific concerns expressed by each group, and explore how these concerns vary across different age ranges. To identify posts that mention specific ages, we conducted a keyword-based search. Table~\ref{tab:age_gender_labels} lists the keyword phrases used. We applied strict phrase matching (e.g., avoiding phrases like ``I have had epilepsy for 5 years”) to ensure higher accuracy, even if it meant excluding some valid data. To assess the statistical significance of the age groups, we used the top $K=10$ TF-IDF features and conducted Chi-square and pairwise t-tests. Following this, we applied Latent Dirichlet Allocation (LDA) topic modeling with $K=3$ topics for each age group to uncover the themes and concerns discussed at different stages. This approach provided insights into how individuals with epilepsy discuss their experiences and challenges based on their age.


\paragraph{Gender of PWE}
A study ~\cite{Hu2021} claimed that men are at a higher risk of developing epilepsy during their lifetime. In our study, we explored this gender dynamics in the discussion of epilepsy. We classified posts as either ``PWE is a Male" or ``PWE is a Female" using keyword based extraction. Table~\ref{tab:age_gender_labels} mentions such gender specific terms we used. To statistically assess these gender-based groups, we performed a Chi-square test using the top $K=50$ TF-IDF features from posts mentioning male and female patients. Additionally, a t-test was conducted on the same TF-IDF features to further explore differences between the groups. Finally, we applied topic modeling using LDA with $K=3$ topics for each gender to identify differences in the content of posts by male and female PWE.

\begin{table}[!htp]
    \centering
    \begin{tabular}{|c|>{\centering\arraybackslash}p{0.6\linewidth}|} \hline 
       Age  & Gender \\ \hline
       18F, 23M, m30  & 18F, 23M, m30 \\ \hline
       22 year(s) old& gender pronouns: she, he, her, his\\ \hline
 I am 29 / He is 36&female: mother, daughter, wife\\\hline
 I turned 34&male: father, son, husband\\\hline
    \end{tabular}
    \caption{Keywords for age \& gender extraction}
    \label{tab:age_gender_labels}
\end{table}
\paragraph{Symptoms and Medications}
To identify key patterns among symptoms and medications mentioned in posts, we first compiled comprehensive lists of symptoms and medications related to epilepsy. The symptom list was obtained from reliable sources such as the NHS~\footnote{\url{https://www.nhs.uk/}} and the Epilepsy Foundation. Using this list, we identified the occurrence of symptoms in posts and comments. A co-occurrence matrix was then calculated to understand relationships between different symptoms. Similarly, for medications, we compiled a list of common antiepileptic drugs (AEDs) from the NHS and the Mayo Clinic, including both generic names and popular brand names. Posts mentioning these medications were identified, and their frequency of occurrence was calculated to analyze the distribution of treatments.

\subsection{RQ2: Temporal Trends}\label{subsec5}
Temporal analysis was conducted to understand community engagement over time. We focused on three key variables: a) the number of posts, b) the average length of posts, and c) the number of comments per post. These variables were selected to provide insights into the level of engagement, the depth of discussions, and the broader trends in awareness within the community. We analyzed the temporal distribution of these variables to observe trends and changes in community interaction and content depth. Figure~\ref{fig:temporalGraph} presents the time-based distribution of these variables.

In addition, we developed a novel engagement metric $F(P)$, defined in Equation~\ref{equation:engagement}, to quantitatively understand user engagement in the subreddit over time. $F(P)$ indicates the overall engagement of a reddit post, reflecting its ability to increase participation in discussions, foster community interaction, and drive meaningful conversations.

The metric is based on the number of comments ($N_{c}$), sentiment scores ($s_i$) of the comments obtained using vaderSentiment~\cite{Hutto2014VADER}, post length ($l$), and readability of the post using the Flesch-Kincaid score ($S_f$). To standardize these variables, we applied max-min normalization, transforming their values to the range $[0,1]$.  Combining these variables into a single engagement metric offers a comprehensive approach to measuring content interaction. This approach integrates quantitative and qualitative aspects: $l$ reflects content depth and detail, influencing reader investment; $s_i$ gauges overall reader sentiment, indicating content reception; and $S_f$ assesses readability, affecting accessibility and audience reach. This method provides a nuanced understanding of how effectively content resonates with its audience, making it a robust measure for assessing content impact and audience engagement levels. 

\begin{equation}\label{equation:engagement}
F(P) = l *  (1 + \sum_{i=1}^{N_c} s_i )  * S_f    
\end{equation}

In this equation, the addition of 1 to $\sum_{i=1}^{N_c} s_i$ ensures that the score $F(P)$ does not become zero even in the absence of comments. This modification allows the score to maintain a meaningful value based on other parameters, ensuring robustness and relevance in evaluating engagement. 

In Table ~\ref{table:correlationfp}, we present the correlation coefficients of each contributing factor with $F(P)$. The correlation analysis shows that the sum of sentiment scores, number of comments, and post length have moderate to strong correlations with $F(P)$, suggesting that these factors significantly influence user engagement. Flesch-Kincaid Score also shows lower positive correlation to $F(P)$ showing lesser role in engagement when compared to others.
\begin{table}[!htp]
    \centering
    \begin{tabular}{lc}
        \toprule
        \textbf{Feature} & \textbf{Correlation with F(P)} \\
        \midrule
        Post Length          & 0.5077 \\
        Number of Comments              & 0.6290 \\
        Sum of Sentiment scores                & 0.5316 \\
        Flesch-Kincaid score & 0.2376 \\
        \bottomrule
    \end{tabular}
    \captionsetup{font=footnotesize}
    \caption{Correlations of $F(P)$ with all contributing factors.}
    \label{table:correlationfp}
\end{table}

\subsection{RQ3: Early Depression Signals}\label{subsec6}
Research indicates that depression and anxiety are significantly more prevalent among people with epilepsy (PWE), with approximately one-third of individuals affected by depression \cite{depression2014kwon}. This highlights the importance of early detection and integrated treatment to enhance their quality of life. In this study, we aimed to detect early signs of depression in Reddit posts and analyze these posts across various demographic categories, symptom patterns, and levels of post engagement.

We utilized the DepRoBERTa-large-depression model \cite{poswiata-perelkiewicz-2022-opi} to evaluate depression levels in Reddit posts. The DepRoBERTa model is a transformer-based language model built on RoBERTa and fine-tuned on a large corpus of Reddit posts related to mental health topics, including depression, anxiety, and suicide. The model classified the posts as severely depressed (0), moderately depressed (1), or not depressed (2). The model's accuracy, as reported in \cite{poswiata-perelkiewicz-2022-opi}, is 62.6\% on the test data. Since the model was already fine-tuned on mental health data from Reddit \cite{sampath2022data}, we used it in a zero-shot setting without additional retraining on our dataset.

To validate the model on our dataset, two annotators manually labeled a random sample of 360 posts. The annotators followed same instructions as those in the original study \cite{sampath2022data}. The initial inter-annotator agreement as measured by Cohen’s Kappa ($\kappa$) \cite{cohen1960coefficient}, was 0.522 (moderate agreement). Disagreements were then handled using mutual discussion. When evaluated on this manually labeled sample, the model achieved an accuracy of 60.6\%. A notable trend was that the model over classified posts labeled as ``moderately depressed" into the ``severely depressed" category, particularly when the posts mentioned multiple treatments or medications, even if the overall tone did not indicate severe depression. 

\section{Results}\label{sec5}
\subsection{RQ1: Demographic Analysis} \label{subsec7}

\paragraph{Who is the PWE?}
The categorization technique revealed that the majority of posts (75.39\%) in the subreddit are authored by individuals with epilepsy (self), while the remaining 22.3\% are contributed by their caretakers. Among caretakers, posts for children with epilepsy are found 8.97\%, posts for romantic partners with epilepsy are found 6.91\% of the times, followed by parents with epilepsy are found 6.44\% of the times. These figures highlight the active involvement of both people with epilepsy and their support networks in the community.

The significance tests between relationship categories yielded a chi-square statistic of 3961.454 (p-value = 1.539e-05), indicating significant differences between groups. The t-tests further revealed significant differences in several pairwise comparisons (see Table~\ref{tab:ttestrelatiop}), though no significant differences were observed between the "Self vs. Romantic" and "Parent vs. Romantic" categories ($p > 0.05$). This suggests that while there are distinct concerns between caregivers and PWE, romantic relationships may share similar emotional and practical challenges with those of parents and PWEs.

\begin{table}[ht]
\centering
\begin{tabular}{|l|c|c|}
\hline
\textbf{Comparison Categories} & \textbf{$t$-Statistic} & \textbf{$p$-Value} \\ \hline
Self vs. Parent & -2.8036 & 0.0051 \\ \hline
Self vs. Children & -2.2048 & 0.0275 \\ \hline
Self vs. Romantic & 1.5619 & 0.1184 \\ \hline
Parent vs. Romantic & -1.3716 & 0.1703 \\ \hline
Children vs. Romantic & -3.4592 & 0.0006 \\ \hline
Parent vs. Children & -2.9992 & 0.0027 \\ \hline
\end{tabular}
\captionsetup{font=footnotesize}
\caption{Pairwise $t$-Test Results between top $K=50$ TF-IDF distributions of different relationship groups.}
\label{tab:ttestrelatiop}
\end{table}

Topic modeling ($K=3, \lambda=0.5$) revealed some common themes across all relationship categories, particularly surrounding seizure episodes, medical health, and the emotional toll of living with epilepsy. For instance, topics such as memory issues, cognitive difficulties, and concerns with medication were consistently observed across the relationship contexts of self, children, parents, and romantic partners. The frequent occurrence of terms like ``brain," ``sleep," ``medication," and ``neuro" highlights the shared concerns about health management, regardless of the relationship type.

However, distinct themes also appeared within each relationship context. In self-related discussions, individuals often expressed feelings of isolation and the need for support, with terms such as ``alone," ``help," ``need," ``job," and ``drive" reflecting personal struggles in balancing daily life with epilepsy. In romantic relationships, there was a noticeable focus on how epilepsy affects intimacy and relationship dynamics, as evidenced by words like ``sex," ``pregnancy," ``love," ``friends," and ``dating." Posts in the children category highlighted concerns about parenting and education, with common terms such as ``school," ``child," ``license," ``college," and ``advice" pointing to issues around managing epilepsy while raising children. In contrast, discussions in the parents category frequently revolved around more severe symptoms and caregiving responsibilities, with terms like ``overshadowing," ``employment," ``jerking," ``wet," ``yelling," ``screaming," ``fallen," ``conscious," and ``smell" emphasizing the practical and emotional challenges faced by those caring for children with epilepsy or managing epilepsy in a family context. 

\begin{figure}[!tp]
    \centering
    \includegraphics[width=0.8\linewidth]{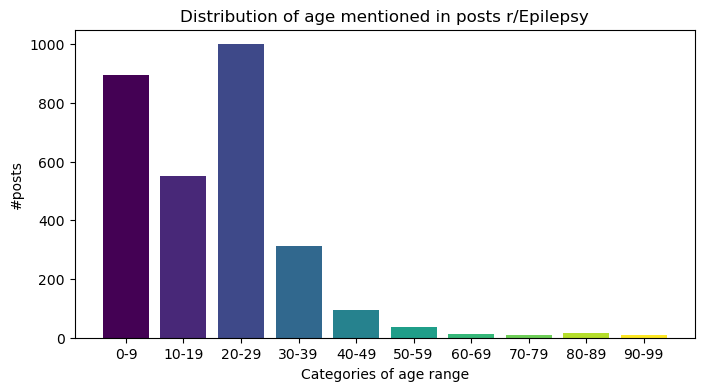}
    \caption{Distribution of age mentioned in r/Epilepsy posts.}
    \label{fig:age-graph}
\end{figure}

\paragraph{Average Age of PWE}
Keyword-based search filtered 2,935 posts, with the highest frequency of mentions in the 20-29 age group, followed by the 0-10 age group (see Figure~\ref{fig:age-graph}). A Chi-Square test between $K=10$ TF-IDF yielded a statistic of 41.07 ($df = 18$, $p = 0.0015$), indicating significant differences in language use across age groups. Pairwise t-tests (Table~\ref{tab:age-pairwise-t-tests}) also showed significant differences between all age group comparisons.


\begin{table}[ht]
\centering
\begin{tabular}{|l|c|c|}
\hline
Comparison (Age Groups) & $t$-statistic & $p$-value   \\ \hline
0-20 vs. 21-40          & $-4.19$       & $2.98e-05$  \\ \hline
0-20 vs. 41-100         & $-4.14$       & $3.77e-05$  \\ \hline
21-40 vs. 41-100        & $-2.18$       & $0.02996$   \\ \hline
\end{tabular}
\captionsetup{font=footnotesize}
\caption{Pairwise $t$-Test Results between top $K=50$ TF-IDF distributions of different age groups.}
\label{tab:age-pairwise-t-tests}
\end{table}

Topic modeling for the 0-20 age group revealed three key themes. The most prominent topic (82.7\% of the discussion) centered on parents and teenagers seeking advice and sharing experiences about epilepsy, with sleep issues frequently mentioned. The second topic focused on managing school life, with terms like ``school," ``semester," and ``absent" reflecting concerns about missing classes, grades, and overall stress due to epilepsy. The third topic involved discussions around gaming, with words like ``VR," ``screen," and ``graphic" highlighting concerns about photosensitivity~\cite{da2017photosensitivity, wolf1986relation} and the potential for seizures triggered by gaming. Additionally, ``marijuana" was a recurring term, indicating curiosity among teenagers, suggesting the need for responsible information about its role in epilepsy management.

For the 20-40 age group, topic modeling uncovered three key themes. The first, covering 40.2\% of tokens, involved balancing work, social life, and epilepsy, with concerns about employment and driving restrictions, indicated by terms like ``job", ``family," and ``license." The second theme, with terms like ``pregnancy," ``surgery," and ``MRI," reflected concerns about epilepsy’s impact on pregnancy and the health of both the mother and baby. The third theme feature terms such as ``feeling," ``anxiety," and ``panic,", focusing on the emotional and cognitive effects of epilepsy, emphasizing anxiety, memory problems, and sleep disturbances.


For the 40-100 age group, three key themes emerged. The first highlighted continued concerns about work and driving, similar to the 20-40 group, with terms like ``job", ``drive" and ``license." The second topic included terms such as ``memory," ``sleep," ``think," ``head" and ``dementia," pointing to cognitive problems such as memory impairment and sleep disturbances, which are commonly studied in relation to epilepsy \cite{helmstaedter2001memory, butler2008recent}. The presence of terms related to dementia in this older age group suggests possible cognitive decline alongside epilepsy, offering a valuable direction for future research \cite{hommet2008epilepsy, sen2018cognition}.The third theme focused on physical symptoms, including sensory disturbances and involuntary movements, indicated by words such as ``ear," ``nose," and ``jerk."

\paragraph{Gender of PWE}
A total of 8374 posts (14.5\% of the dataset) were identified as containing gender-related terms. Out of the posts that mentioned gender, 54.78\% referred to male patients, indicating a slight male bias in the dataset. The chi-square test comparing the distribution of posts mentioning male versus female patients yielded a chi-square statistic of 1119.542, with a $p$-value of 0.0045 and degrees of freedom of 999. This low $p$-value suggests a significant variation in the way epilepsy is discussed with respect to different genders. Additionally, a t-test on the top $K=50$ TF-IDF features revealed a t-statistic of 2.3455 and a $p$-value of 0.0190, indicating a statistically significant difference in how male and female patients are mentioned in posts.

Further, topic modeling with $\lambda=0.5$ revealed general themes across both male and female posts, including discussions on symptoms, sleep disturbances, and medical health. To investigate potential gender-specific differences, we lowered $\lambda$ to 0.3, which allowed more distinct topics to emerge. 

For male PWE posts, the dominant topics included discussions around seizure episodes, with a focus on physical experiences like jerking movements, confusion, and memory issues. Words like ``episode," ``panic," and ``focal" were prominent, suggesting that men with epilepsy may be more concerned with describing their physical seizure experiences and the immediate aftermath, including confusion and cognitive impacts.  Additionally, topics related to work, insurance, and support were prevalent, reflecting practical concerns around managing epilepsy in everyday life. For female PWE posts, a noticeable shift was observed toward discussions around pregnancy and hormonal influences, with terms such as ``pregnancy," ``hormonal," ``risk," ``newborn" and ``menstrual" being key indicators. This suggests that women expressed concerns about the impact of epilepsy and its treatments on pregnancy and hormonal cycles. Furthermore, mental health and emotional well-being were significant themes, with frequent mentions of anxiety, panic attacks, and sleep disturbances. 

\begin{figure}[!htp]
\centering
\includegraphics[width=0.6\linewidth]{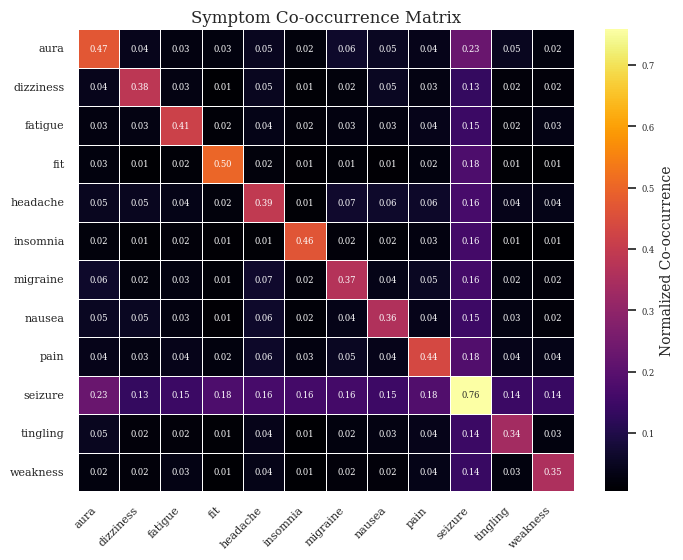}
\caption{Correlation Matrix of Symptoms in Epilepsy.}
\label{fig:symptomCoocurrence}
\end{figure}

\paragraph{Symptoms and Medications}
Figure~\ref{fig:symptomCoocurrence} represents the co-occurrence matrix of different symptoms experienced during epilepsy as mentioned in the posts with the term ``seizure" mentioned most, with its strongest co-occurrence observed with ``aura" (0.23), indicating that auras, often serving as a precursor to seizures, are commonly discussed in the community. Seizures also showed notable correlations with ``headaches" (0.18) and ``insomnia" (0.16), highlighting these as frequent issues among individuals with epilepsy.

\begin{figure}[!ht]
    \centering
    \includegraphics[width=1\linewidth]{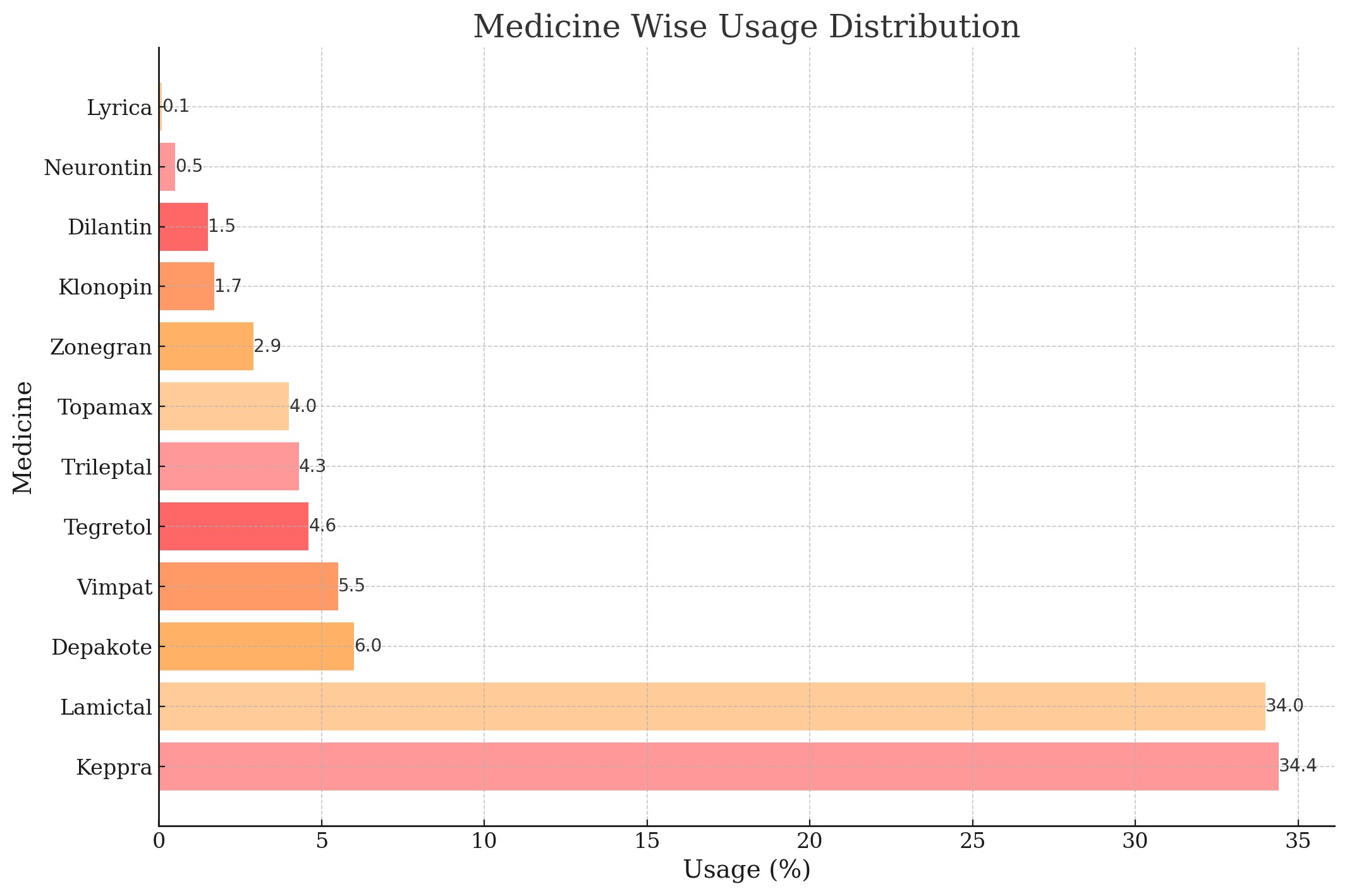}
    \caption{Distribution of Medications in r/Epilepsy posts.}
    \label{fig:medication}
\end{figure}

Figure~\ref{fig:medication} shows different commonly discussed medications in epilepsy posts. Keppra\footnote{\url{https://www.nhs.uk/medicines/levetiracetam/}} was identified as the most common antiepileptic drug, mentioned in 34.4\% of posts, closely followed by Lamictal\footnote{\url{https://www.nhs.uk/medicines/Lamotrigine}} (34.0\%). Both are widely used for managing a broad range of seizures, including partial-onset seizures and generalized tonic-clonic seizures.

\subsection{RQ2: Temporal Trends}\label{subsec8}
The temporal analysis revealed a dynamic and growing community, as shown in Figure~\ref{fig:temporalGraph}. The number of daily posts increased by approximately 20.53\% each year, reflecting rising engagement within the subreddit. This growth was accompanied by a 4.11\% annual increase in the average post length, indicating that users are sharing more detailed content, such as personal stories and in-depth discussions, which enrich the community environment. Additionally, the number of comments per post rose by 2.86\% annually, further highlighting the increase in community interaction.

\begin{figure}[!htp]
    \centering
    \includegraphics[width=\linewidth]{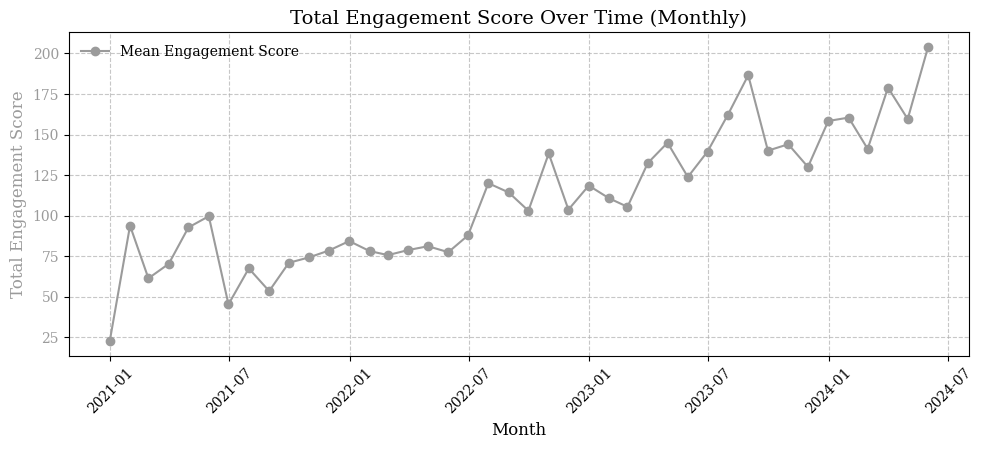}
    \caption{Temporal trend of engagement scores.}
    \label{fig:engagement}
\end{figure}

We also evaluated our engagement metric $F(P)$ over time. As shown in Figure~\ref{fig:engagement}, the engagement score exhibited an 11.79\% month-on-month increase, suggesting that interactions within the community are becoming not only more frequent but also more meaningful. Users are contributing longer, more detailed discussions, sharing advice, fostering deeper connections, and exchanging personal insights and advice. This upward trend in community engagement serve as important indicators of a healthy, growing online space. 

Awareness initiatives, such as International Epilepsy Day (held on the second Monday of February) and Purple Day (on March 26), play a crucial role in raising awareness about epilepsy, debunking misinformation, and promoting advocacy for people with epilepsy (PWE).

A recent study \cite{mugumbate2023epilepsy} raised concerns about the cost-effectiveness of such epilepsy awareness initiatives and the potential limitations posed by regional and cultural differences. However, we believe that social media communities like \texttt{r/Epilepsy} can help overcome these challenges by serving as effective platforms for disseminating accurate information, providing a safe space to share personal stories, and offering mutual support.

To further explore the role of online communities in supporting awareness efforts, we analyzed the activity around key epilepsy awareness days, including International Epilepsy Day and Purple Day, as marked in the Figure~\ref{fig:temporalGraph}. Overall, we observed an average increase of more than 8\% in the number of posts around these specific dates compared to the preceding 30 days. Specifically, in 2024, 70 posts were authored on 2nd Monday on Feburary, and 69 posts on 26th March (nearly 1.75 times the average). We see an average 8\% posts mentioning phrases like ``World Epilepsy Day" and ``Purple Day" on these days. This suggests that these awareness initiatives might be effective at generating discussions and engagement within the subreddit. The significant increase in activity around these key dates reinforces our claim that online communities are powerful tools for spreading awareness and fostering engagement, especially in global and culturally diverse contexts.
\subsection{RQ3: Early Depression Signals}\label{subsec9}
The DepRoBERTa model classified approximately 39.75\% of the posts with ``severe depression" signals and about 9.43\% with ``moderate depression" signals. These depression signals were then analyzed across different categories, including relationships, age, and gender mentioned in the posts as well as common symptoms and treatments. The analysis revealed various insights discussed below.

\begin{figure}[!htp]
    \centering
    \includegraphics[width=1\linewidth]{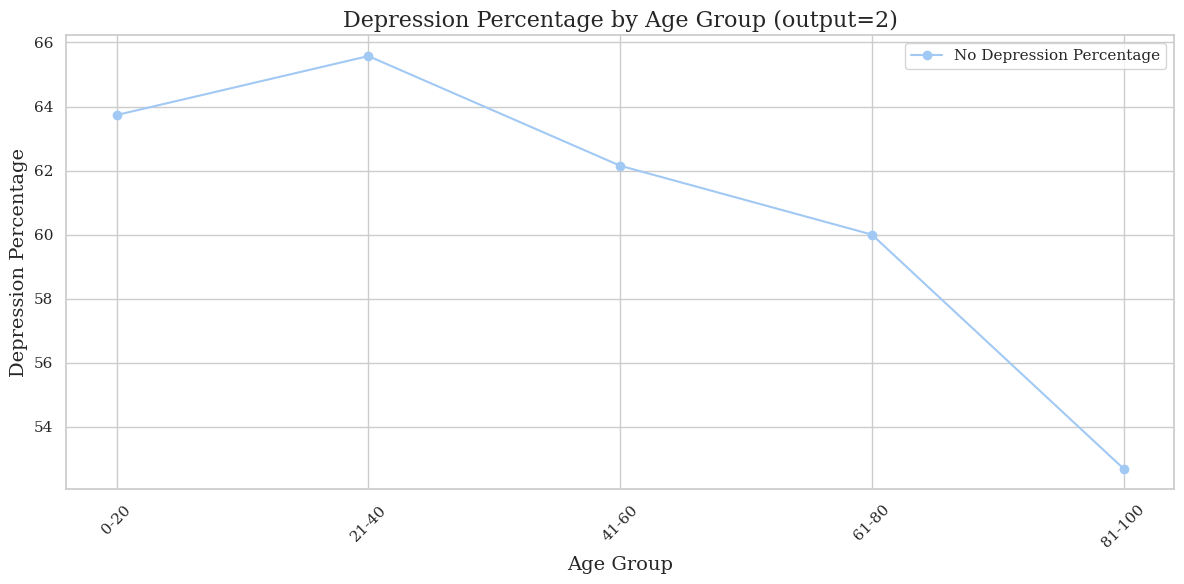}
    \caption{Depression percentage variation with age.}
    \label{fig:ageDepression}
\end{figure}

Firstly, we analysed the depression signals across different age categories. As illustrated in Figure~\ref{fig:ageDepression}, a consistent trend of reduced depression signals with increasing age was observed. Notably, our analysis identified a peak in depression signals among individuals aged 21-40 years. This aligns with \cite{grabowska2006risk}, which reports a relatively low frequency of depression in individuals over 40 in PWE. To assess the statistical significance of this trend, we divided our dataset into 5-year intervals and conducted a linear regression analysis, using the midpoint of each age interval as the predictor variable and the corresponding depression signals as the outcome. The regression analysis yielded a significant $p$-value of 0.012 for the age midpoint coefficient, confirming an inverse relationship between age and depression signals. This suggests that younger age groups, particularly those between 21 and 40, exhibit higher levels of depression compared to older adults. This highlights the need for targeted intervention and support strategies for PWE in younger and middle aged groups.


Next, we found no statistically significant differences in depression levels based on gender categories mentioned in the posts ($p > 0.05$), so no further investigation was conducted. This aligns with findings from \cite{liu2020gender}, which observed similar levels of depression and anxiety in men and women in a study of 158 PWE. However, other studies \cite{gaus2015gender, burkert2015gender} report higher depression levels in females, suggesting that a more thorough evaluation may be necessary.

\begin{figure}[!htp]
    \centering
    \includegraphics[width=0.6\linewidth]{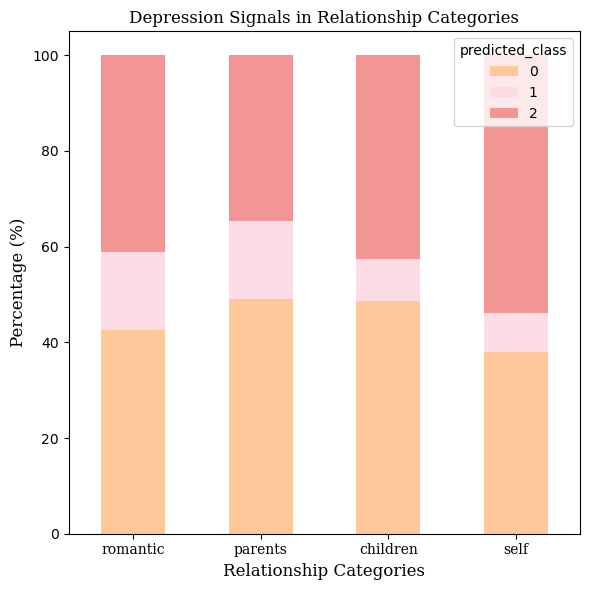}
    \caption{Depression percentage in different relationship.}
    \label{fig:relationDepression}
\end{figure}

Lastly, depression signals were analysed between different types of relationships mentioned. Figure \ref{fig:relationDepression} presents the percentage of severity levels across romantic, parents, children and self categories. 

Our analysis reveals notable differences in the intensity of depression expressed across various relationship categories. Posts categorized under ``self" show a lower incidence of severe depression (~38\%) compared to ``romantic" (~43\%), ``parents" (~48\%), and ``children" (~47\%). This suggests that individuals who write about their own experiences may, in general, report less severe depression, possibly due to a lower level of emotional intensity or a tendency to minimize their own distress. In contrast, expressions of concern for others, particularly loved ones, appear more intense. In ``parents" category, over 65\% of posts were classified as depressed (0 and 1), indicating a significant emotional distress related to the well-being of loved ones. Statistical analysis confirms that these differences are statistically significant($p = 0.0419$), implying that concerns for others provoke stronger emotional responses than self-reports. This highlights the influence of relational dynamics on mental health discussions.

\begin{figure}[!htp]
    \centering
    \includegraphics[width=1\linewidth]{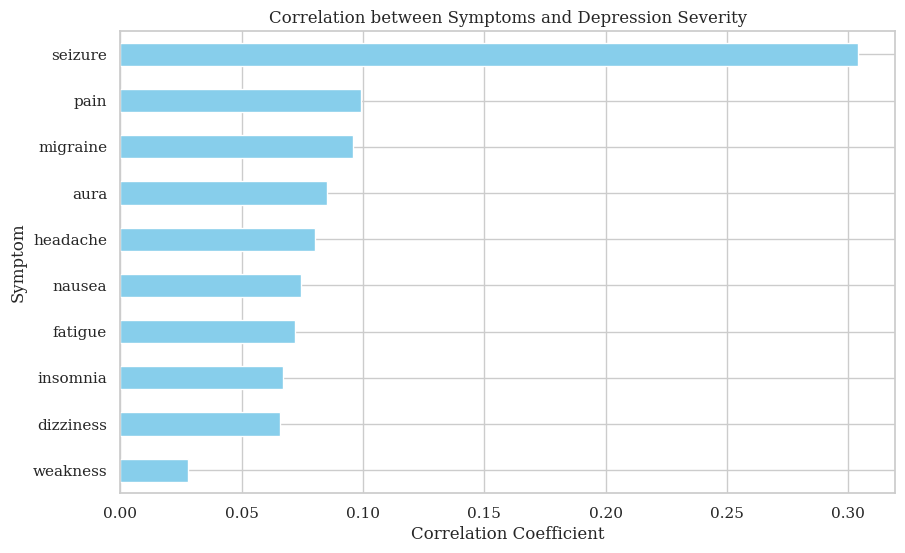}
    \caption{Symptoms correlation to depression severity.}
    \label{fig:depressioncor}
\end{figure}

Apart from demographic based depression analysis, we also examined the correlation between the occurrence of common symptoms and depression levels. The Pearson correlation coefficient, as presented in Figure~\ref{fig:depressioncor}, revealed seizures as the most strongly correlating symptom, with a correlation value exceeding 31\%. This significant correlation underscores the profound emotional impact of recurrent seizures on mental health. Further, given the frequent co-occurrence of seizures with other symptoms like headaches, migraines, and nausea, we also observed positive correlations, suggesting broader neurological implications.

Finally, we also investigated the community's response to evident depression signals. Specifically, we analyzed the correlation between the engagement score $F(P)$, as defined in Equation~\ref{equation:engagement}, and the depression signals using Pearson correlation coefficient. Our results demonstrated a strong correlation of 28.6\% indicating that higher engagement scores are observed when the posts exhibits evident depression signals. This indicates that the community actively supports individuals experiencing challenges and facing mental health issues. These findings highlight the positive impact of community support and the benefits people can obtain from online social help groups such as Reddit.
\section{Key Insights and Discussions}\label{sec8}
\paragraph{Social Media Communities as a Tool for Outreach}Our analysis showed that \texttt{r/Epilepsy} is a growing community with increased activity during epilepsy awareness dates and higher engagement when depression signals are present. This highlights the potential of social media communities for outreach, providing real-time opportunities to engage with PWE, share crucial information, raise awareness, and offer timely intervention and support.

\paragraph{Workplace Safety and Support for PWE} Work-related concerns were common across different age groups, relationships, and genders. A closer manual analysis of 50 filtered posts highlighted PWE's worries about workplace seizures, job security, and safety. More open workplace policies are needed, including accommodations, epilepsy education, and seizure management plans. 

\paragraph{Driving Regulations for PWE} PWE expressed concerns about the emotional \& logistical challenges of losing driving privileges due to seizures. While necessary for public safety, better communication and support systems are needed to help manage transportation limitations. Providing alternative transport options and policies that promote independence will help ease these burdens, especially in countries requiring seizure-free periods to drive \cite{driving2023moller}.

\paragraph{Medicinal Marijuana for Teenagers with Epilepsy} Our study found that marijuana was frequently discussed by PWE aged 0-20, suggesting curiosity or temptation among teenagers. With research on medicinal marijuana and epilepsy growing \cite{maa2014case, ladino2014medicinal, gordon2001alcohol}, it is crucial to provide clear, responsible information to prevent misuse and address misconceptions about its role in epilepsy management.

\paragraph{Support for Teen Gamers with Epilepsy} Another concern among teenagers that emerged in our study was photosensitivity, with seizures triggered by gaming. This highlights the need for gaming companies to develop more seizure-friendly games, allowing teens to enjoy gaming while managing their health.

\paragraph{Need for Gender-Specific Outreach in Epilepsy} Our study revealed that women with epilepsy frequently discuss reproductive health, hormonal changes, and pregnancy. While some research exists \cite{hophing2022sex,tauboll2008gender}, there is a significant gap in outreach efforts. More accessible resources are needed to address how epilepsy and medications interact with hormonal cycles and pregnancy, providing women with proper guidance and support.

\paragraph{Support for Parents of Children with Epilepsy} Parents expressed concerns about their children's education and social development, citing inadequate school support systems. Concerns regarding missing school, academic struggles and social isolation. There is a need for better school policies, including seizure management plans, teacher training, and accommodations like flexible schedules.

\paragraph{Mental Health and Epilepsy Co-Management}
Our depression signal analysis revealed that epilepsy often co-occurs with anxiety and depression, underscoring the need for integrated care. Current treatments may overlook the emotional burden, including isolation, fear of seizures, and cognitive issues. Involving mental health professionals in epilepsy care and encouraging PWE to seek routine mental health support is essential.

\section{Potential Biases and Limitations}\label{sec9}
Our study has several limitations in extracting data from Reddit's r/Epilepsy community. The limited availability of demographic information introduces selection bias, as the characteristics we obtained may not represent the broader epilepsy population. Additionally, many users remain anonymous, leading to gaps in key data such as age, gender, and socioeconomic background, which could result in response bias—those sharing personal details may represent more extreme experiences. The DepRoBERTa model used in our analysis has an accuracy of 60.6\%, potentially affecting the reliability of depression signal results. 

\section{Ethics Statement}\label{sec10}
In this study, we use publicly available data from Reddit, where users post anonymously. Although we include a few direct quotes, all personally identifiable information (PII) has been removed to ensure anonymity. Reddit’s anonymous nature reduces identification risks, but ethical concerns remain. Sharing insights on sensitive topics like depression and epilepsy could cause anxiety among users, and there is a risk of misuse in contexts like recruitment or insurance. We stress that this research is intended solely to advance understanding and support for individuals with epilepsy, not for exploitation or discrimination.

\section{Conclusion}\label{sec11}

Our study highlights the valuable role of social media platforms in understanding the challenges faced by people with epilepsy. Key themes include workplace safety, driving regulations, mental health, and gender-specific outreach. However, limitations like selection bias and anonymized data require caution when generalizing findings. Future research should address these limitations and enhance support strategies for diverse epilepsy populations.

\section*{Declarations}
\begin{itemize}
\item \textbf{Funding: }No funding was received for conducting this study.
\item \textbf{Competing Interests: }The authors have no competing interests to declare that are relevant to the content of this article.
\item \textbf{Ethics Statement: } Please refer \ref{sec10}
\item \textbf{Consent for publication: }No formal consent was required as the data is publicly available, and both authors consent to the publication of this work.
\item \textbf{Code and Data: } Both dataset prepared and code has been made available at \url{https://shorturl.at/neGz7} (to be replaced by GitHub link on acceptance). 
\item \textbf{Author contribution: }Both authors have equal contribution.
\end{itemize}

\end{document}